\documentclass{article}
\pdfoutput=1
\usepackage[a4paper, total={6in, 10in}]{geometry}

\usepackage{authblk}
\usepackage{caption}
\usepackage{graphicx}
\usepackage{bm}
\usepackage{bbm}
\usepackage{amsmath}
\usepackage{amssymb}
\usepackage{comment}
\usepackage{wrapfig}
\usepackage{xcolor}
\usepackage{times}
\usepackage[numbers]{natbib}

\newcommand\independent{\protect\mathpalette{\protect\independenT}{\perp}}
\def\independenT#1#2{\mathrel{\rlap{$#1#2$}\mkern2mu{#1#2}}}

\title{Causal Effect Variational Autoencoder with Uniform Treatment}

\author[1]{Daniel Jiwoong Im}
\author[1,3]{Kyunghyun Cho}
\author[2]{Narges Razavian}

\affil[1]{New York University, CDS}
\affil[2]{NYU Langone Health}
\affil[3]{Genentech}
\date{} 

\begin{document}

\maketitle

\begin{abstract}%
    Domain adaptation and covariate shift are big issues in deep learning and they ultimately affect any causal inference algorithms that rely on deep neural networks. Causal effect variational autoencoder (CEVAE) is trained to predict the outcome given observational treatment data and it suffers from the distribution shift at test time. In this paper, we introduce uniform treatment variational autoencoders (UTVAE) that are trained with uniform treatment distribution using importance sampling and show that using uniform treatment over observational treatment distribution leads to better causal inference by mitigating the distribution shift that occurs from training to test time. We also explore the combination of uniform and observational treatment distributions with inference and generative network training objectives to find a better training procedure for inferring treatment effect. Experimentally, we find that the proposed UTVAE yields better absolute average treatment effect error and precision in the estimation of heterogeneous effect error than the CEVAE on synthetic and IHDP datasets.
\end{abstract}


\section{Introduction}

Inferring causal relationships is one of the fundamental problems in science, medicine, and many other application domains \citep{Pearl2010, Yazdani2015, LaLonde1986, Banerjee2012, Stovitz109, Raita2021}. 
The problem of causal inference is often regarded as asking counterfactual questions, because being able to infer these counterfactual outcomes helps us understand the effect of an intervention on
an individual who is often represented as a set of covariates. For example, what would have happened if the Covid-19 patient took a particular medication? Which of the existing Covid-19 treatments work the best for a given patient?
Having a machine learning method that can answer such  questions using observational data creates powerful tools for realizing the nature of how things work \citep{Pearl2009}.
Our aim is to learn a model that can infer the counterfactual outcomes and use them as evidence to understand the causal effect using observational data.

One of the challenges of measuring causal effect from observational data is that the assignment of a treatment (i.e prescription of a medication) does not happen at random and depends on a set of confounders, which in-tern influence the outcome. 
\citep{Pearl2015, Wager2015, Peysakhovich2016, Johansson2016}.
There are various proposed methods to solve this challenge, such as front door criterion \citep{Pearl1993}, instrumental variables \citep{Angrist1996, Hernan2006, Imbens2014},
causal structure learning \citep{Frot2017, Shi2019}, and more \citep{Kuroki2014, Johansson2016, Rothenhausler2018, Wang2019}. In this paper, we focus on a setting where the confounder is unknown - as 
it is unrealistic to measure all possible confounders in practice- and yet, we have proxy variables that are observable covariates representing the noisy version of the confounders \citep{Pearl2021,Kuroki2014}.
For example, let Covid-19 medication be a treatment $T$, recovery result be an outcome $Y$, unknown confounder be the latent variable $Z$, and the patient's electronic health records (EHR) data be the observation $X$. 
The hope is to recover the hidden confounder $Z$ with the help of the patient's EHR data and infer the treatment outcome using the confounder $Z$.

The variational autoencoder (VAE) has been applied to train latent variable based deep generative models, which estimates an observation distribution $p(X)$ by modelling latent variable $Z$ \citep{Kingma2014}.
A VAE is trained using approximate posterior samples from a simultaneously trained inference network to maximize the evidence lower bound.
\cite{Louizos2017} proposed a causal effect variational autoencoder (CEVAE) that adapts VAE for causal inference problems.
The causal graph of CEVAE in Figure~\ref{figs:graph_structure}b represents the generative process by which observational samples are drawn. 

The advantage of CEVAE is that it only requires observational data that is collected without systematic interventions \citep{Colnet2021}. 
Once trained on such observational data, we can use CEVAE to estimate a treatment effect with a treatment distribution that is independent of the confounding factors.
However, this inference-time treatment distribution is not a treatment distribution from which the treatments were drawn in the observational data.
In other words, there is discrepancy between training the CEVAE and using it for causal inference, which is serious issue in practice due to potentially significant distribution shift.

A divergence of the model output distribution due to treatment distribution shift is amplified especially when the causal inference model relies on deep learning. 
It is well studied that deep neural networks performance changes dramatically when the distribution shifts
\cite{Sergey2015, Recht2019, Yadav2019, Shimodaira2000, Jeong2020RobustCI, Lazaridou2021}. Consequently the performance of causal inference model can differ significantly even when the posivity condition is satisfied in the covariate space. We therefore investigate a way to train a CEVAE to minimize this discrepancy.


In this paper, we propose to train a CEVAE on data where the treatment variable is conditionally independent of confounding factors. 
Such data is not available in reality and we only have access to the observational data where the treatment variable depends on confounding factors. 
We use importance sampling to approximate the proposed objective function with the observational data to ensure uniform target treatment distribution during the training.
In order to make efficient use of observational and uniform treatment distributions, we explore combining the two distributions while training inference and generative networks.
We experiment with (i) training both inference and generative networks with only observational data, (ii) training both inference and generative networks using only uniform treatment distribution, (iii) training inference and generative network with uniform and observational data respectively, and (iv) visa versa. 
In the experiments, we empirically demonstrate that the proposed method improves the performance in both a synthetic and IHDP dataset in terms of recovering average treatment effects.

\section{Background}
\subsection{Individual Treatment Effect}
\begin{figure}[t!]
\centering
\includegraphics[width=\linewidth, height=0.25\textheight]{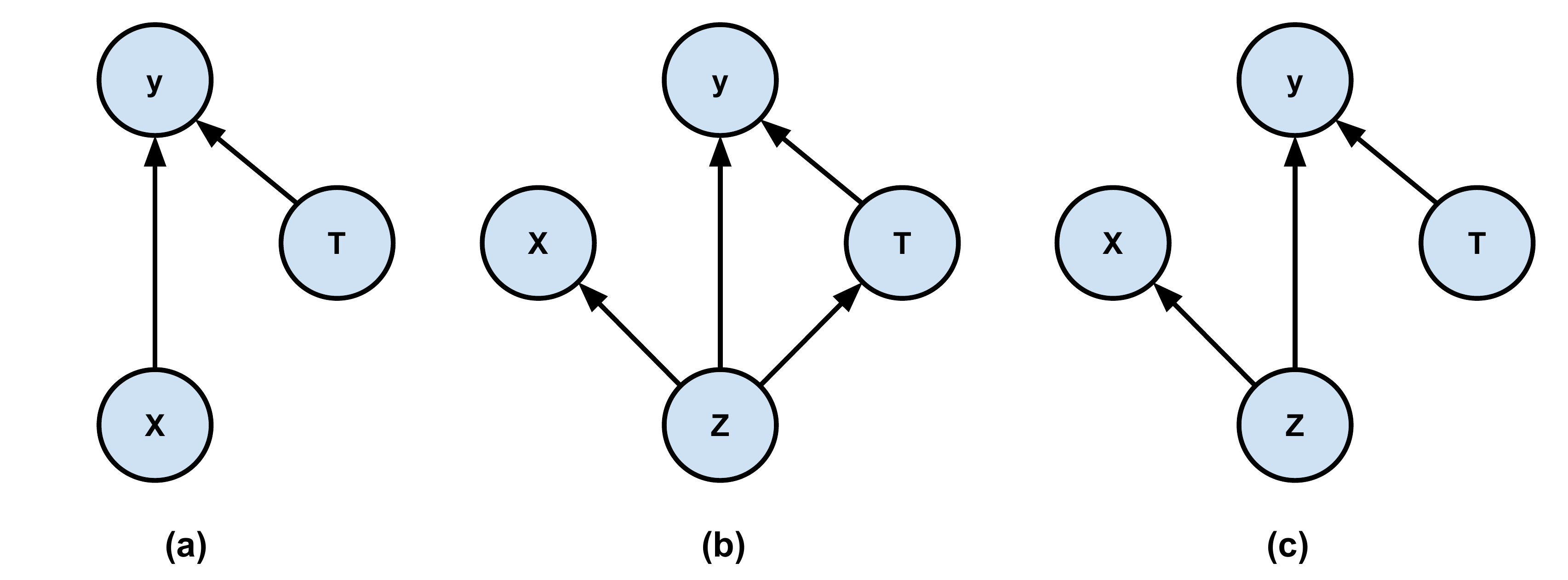}
\caption{Graphical Structures - (a) A naive causal graph where the observable confounders $X$ together with the independent treatment $T$ determines the outcome $Y$. (b) A causal graph that often corresponds to the data generation process in real-life where an unobservable confounder $Z$ influences the treatment decision $T$, and $X$ is an observable proxy for confounder $Z$.  (c) A causal graph that corresponds to a probabilistic graphical model that is trained using uniform treatment distribution.}
\label{figs:graph_structure}
\end{figure}

In this paper, we work with graphical structures that specifies the 
causal connections between observable proxy $X$, unobserved confounder $Z$, treatment $T$, and outcome $Y$ as shown in Figure~\ref{figs:graph_structure}. 

We are interested in a binary treatment assignment setting where $1$ indicates the treated action and $0$ indicates the control action. 
We assume that consistency assumption holds where $Y=Y(1)T + Y(0)(1-T)$.
In order to measure the treatment effect, we compute the difference of treated and controlled outcomes for a given patient $Y(1) - Y(0)|X=x$.
Then, the individual-level treatment effect of $X=x$ on $Y$ is defined as
\begin{align*}
    ITE(x) := \mathbb{E}[y|X=x, do(t=1)] - \mathbb{E}[y|X=x,\mathrm{do}(t=0)]
\end{align*}
and the population average treatment effect is $ATE := \mathbb{E}_{p(x)}[ITE(x)]$.
The treatment has no effect on a patient $x$ if both treated and controlled outcomes are the same (i.e., $ITE(x)\simeq0$).
Otherwise, the treatment has either positive or negative effect on an individual patient $x$.

\subsection{Causal Effect Variational autoencoder}
We can infer the causal effect using a probabilistic graphical model with the dependency structure induced by the graph.
In deep latent variable models for causal inference, we assume a particular data generating process described in a probabilistic graphical model from Figure~\ref{figs:graph_structure}b.
In below, we describe how the treatment effect can be computed under different assumptions and how the distribution in this model can be effectively and efficiently estimated.

In this setup, latent variable $Z$ is a hidden confounder, to which we have access to via observable proxy $X$ (see Figure~\ref{figs:graph_structure}b).
We assume that we can recover the joint distribution $p(Z,X,T,Y)$ from observational data $(X,T,Y)$. 
This enables us to compute $p(Y|X=x,do(T=1))$ and $p(Y|X=x,do(T=0))$
which are required to for computing $ITE(x)$:
\begin{align*}
    p(Y|X=x,do(T=t)) &= \int_z p(Y|X,do(T=t),Z) p(Z|X,do(T=t))dZ\\
                    &= \int_z p(Y|X,T=t,Z) p(Z|X)dZ.
\end{align*}
The second equality follows from the rule of {\em do}-calculus and the assumption that
$Z$ is independent of $T$ \citep{Louizos2017}.
Therefore, we can estimate $p(Y|X=x,do(T=1))$ as long as we can approximate $p(Y|X,T=1,Z)$ and $p(Z|X)$.

The casual effect variational autoencoder is a particular type of variational inference framework which allows us to estimate $p(Y|X,T,Z)$ and $p(Z|X)$ using deep neural networks \citep{Kingma2014, Louizos2017}.
With the VAE, the posterior distribution is defined as $p_\theta(Z|X,T,Y) \propto p_\theta(X,T,Y|Z)p(Z)$. 
Specifically, we define a prior $p(Z)$ on the latent variable $Z$.
Then, we use a parameterized distribution to define the observation model $p_\theta(X,Y,T|Z)$
using a neural network with the parameter $\theta$. The input is $z$
and output is a parametric distribution over $(X,Y,T)$, such as the Gaussian or Bernoulli, depending on the
data type. Additionally, we approximate posterior distribution $q_\phi(Z|X,T,Y)$ using a neural network with
variational parameter $\phi$. We infer the hidden confounder $Z$ from the observations $(X,T,Y)$ using a neural network.

We estimate $p(Y|X,T=1,Z)$ and $p(Z|X)$ directly by training both generative and inference networks
from observational data.
The objective of VAE is to maximize the following variational lower bound with respect to the
parameters $\theta$ and $\phi$:
\begin{align*}
    \mathcal{L}_{\text{CEVAE}} = \mathbb{E}_{p(X,T,Y)}\left[\mathbb{E}_{q_\phi(Z|X,T,Y)} \left[ \log p_\theta(X,T,Y|Z)\right] - \mathbb{KL}\left[ q_\phi(Z|X,T,Y) \| p(Z)\right]\right].
\end{align*}
The first term is the reconstruction of observable variables from the inferred confounder $Z$, and the
second term is a regularizer which enforces the approximate posterior to be close to the prior and
maximizes the entropy of the inferred confounder $Z$.
We can jointly update both generative and inference network parameters by applying backpropagation
and using the re-parameterization trick \citep{Rezende2014, Kingma2014}.

Having the ability to recover $(Z,X,T,Y)$ by learning the latent distribution of $Z$,
we can now use the model distribution to efficiently compute a counterfactual query, $p(Y|X=x^*,do(T=1))$ and $p(Y|X=x^*,do(T=0)$ where $x^*$ is a query sample. 
Estimating posterior distribution is required in order to answer a counterfactual query.
We estimate posterior distribution $p(Z|X)$ using our approximate posterior distribution $q_\phi(Z|X,T,Y)$ that takes input $x^*$, $t^*$, $y^*$. 
Because we are only given $x$ of an individual, we train two other deep neural networks, $q_\varphi(T|X)$ and $q_\varphi(Y|X,T)$ to infer $t$ and $y$ from $x^*$.
We maximize the log-likelihood over the network parameters $\varphi$ together with the VAE objective:
\begin{align*}
    \mathcal{J}_{\text{CEVAE}} := \mathcal{L}_{\text{VAE}} + \mathbb{E}_{p(X,T,Y)}\left[(\log q_\varphi(T|X) + \log_\varphi q(Y|X,T)\right].
\end{align*}
where $\varphi$ is the parameters of $q_\varphi(T|X)$ and $q_\varphi(Y|X,T)$.
Finally, we estimate 
\begin{align}
    p(Y|X=x^*,do(T=1))&=\int_z p(Y|X=x^*,T=1,Z) p(Z|X)dZ \\
    &\approx \int_z p_\theta(Y|X,T=1,Z) q_\phi(Z|X=x^*,T=t^*,Y=y^*)dZ
    \label{eqn:cevae_estimate}
\end{align}
where $t^*$ and $y^*$ are inferred from the sample $x^*$.
Both training and inference are efficient and expressible for modelling continuous latent variables for well balanced i.i.d datasets. 

Unfortunately, CEVAE is less effective when it comes to measuring ATE,
especially when the treated and controlled actions are highly disproportionate.
This is because CEVAE is trained from observational data while ATE is computed with the treatment intervention.
In the next section, we expound the issue and propose an alternative method to overcome this limitation.


\section{Training CEVAE with uniform treatment distribution}
It is worth paying attention to how treatment actions are used to computing ITE. 
The outcomes are computed over both binary actions through intervention during the inference.
This is equivalent to generating outcomes from a causal graph that does have an edge between the confounder $Z$ and the treatment $T$.
However according to the CEVAE framework, the model has learned the conditional distributions $p_\theta(X,Y,T|Z)$ and $q_\phi(Z|T,X,Y)$ from treatment samples that are not uniformly distributed.
Given that there is no intervention in the training process, the nature of observational data collection process induces a directed edge from confounder $Z$ to treatment $T$.
Therefore, we have a discrepancy in ATE between learning and inference procedure, which corresponds to the distribution shift or domain shift.
It is well-known that distribution shift is detrimental to neural network predictions \cite{Shimodaira2000, Sergey2015, Jeong2020RobustCI}. 
This makes the discrepancy in ATE to be even further apart as causal inference model uses deep neural networks.

Replacing the observational distribution with uniform treatment distribution provides a fair and randomized treatment samples for training CEVAE, to answer counterfactual questions,
because having a uniform treatment selection process naturally decouples $Z$ and $T$ and sets $p(T|X)=p(T)$ to a uniform distribution, similar to a randomized clinical trial over treatment $T$. 
For this reason, we train a latent variable causal model using uniform treatment distribution.
We hypothesize that our latent variable causal model generalize better by correcting the distribution shift between training and inference.
Here, the observational-data based distribution is $p(X,T,Y) = p(T|X)p(X)p(Y|X,T)$ and uniform treatment distribution is $r(X,T,Y) = r(T|X)p(X)p(Y|X,T)$.

Let us re-express the variational lower bound objective function over uniform treatment distribution:
\begin{align*}
    \mathcal{L}_{\text{UTVAE}} := \mathbb{E}_{r(X,T,Y)}\left[ \mathbb{E}_{q_\phi(Z|X,T,Y)} \left[ \log \frac{p_\theta(X,Y,T|Z)p(Z)}{q_\phi(Z|X,T,Y)} \right] \right].
\end{align*}
Since we do not have actual uniform treatment distribution $r(X,T,Y)$ but only have the observational data $p(X,T,Y)$, we rely on importance sampling procedure to estimate the uniform treatment distribution:
\begin{align*}
    \mathcal{L}_{\text{UTVAE}} &= \mathbb{E}_{p(X,T,Y)}\left[ w(X,T) \mathbb{E}_{q_\phi(Z|X,T,Y)} \left[ \log \frac{p_\theta(X,Y,T|Z)p(Z)}{q_\phi(Z|X,T,Y)} \right] \right]
\end{align*}
where $w(X,T) = \frac{r(T|X)}{p(T|X)} = \frac{1}{2 p(T|X)} $ is the importance weight. Note that $\frac{r(T|X)}{p(T|X)} = \frac{r(X,T,Y)}{p(X,T,Y)}$ and $r(T|X)=r(T)=\frac{1}{2}$ are due to independence between $X$ and $T$ in the causal graph and the uniformly distributed treatment selection procedure. 

Given this approach to obtain CEVAE distribution $p_\theta(Z,X,T,Y)$ that is trained using uniform treatment distribution, we compute a counterfactual query at inference time. 
Again, we train neural networks using maximum log-likelihood in order to infer $t$ and $y$ from out-of-sample query $x^*$.
Together with the UTVAE objective, we train additional deep neural networks $q_\varphi(T|X)$ and $q_\varphi(Y|X,T)$ to infer treatment $t^*$ and outcome $y^*$,
\begin{align*}
    \mathcal{J}_{\text{UTVAE}} := \mathcal{L}_{\text{UTVAE}} + \mathbb{E}_{p(X,T,Y)}\left[(\log q_\varphi(T|X) + \log_\varphi q(Y|X,T)\right].
\end{align*}
where $\varphi$ is the parameters of $q_\varphi(T|X)$ and $q_\varphi(Y|X,T)$.
Finally, we approximate $p(Z|X)$ using our approximate posterior distribution $q_\phi(Z|X=x^*,T=t^*,Y=y^*)$ and compute $p(Y|Z,do(T=1))$ and $p(Y|Z,do(T=0))$.

\subsection{Pairing observational and uniform treatment distribution with generative and inference distributions }
In the CEVAE, there are two conditional distributions that depend on treatment $T$, $p_\theta(Y|T,Z)$ and $q_\phi(Z|T,X,Y)$. 
Both of these distributions can be estimated using samples drawn from a treatment distribution that is either dependent on or independent of the confounding factor.
In doing so, we have the option to use observational data based, or uniform treatment distributions, for estimating generative and inference distributions respectively.
The question is then whether there is a particular combination of treatment and conditional distributions, that results in a CEVAE which is better at inferring a treatment effect.


Our primary goal is to recover the true $p(X,T,Y,Z)$ using maximum likelihood (or variational lower bound). At the same time, we also have to
subsequently learn a good representation of hidden confounders. In our framework, these two objectives can be separated for training an inference and generative network. 
Although it is well-known that uniform treatment distribution is ideal for understanding causal effect of given causal graph \citep{Pearl2009, Aronow2021}, it is unclear
whether it is also suitable for learning a good representation of hidden confounders. 
There are four possible objective functions that we can explore by permuting uniform and observational-data-based distributions with separate inference and generative objective functions. 
We already describe $\mathcal{L}_{\text{CEVAE}}$ and $\mathcal{L}_{\text{UTVAE}}$ which only uses either an observational or uniform distribution. 
Here we list the remaining two objective functions:
\begin{align*}
    \mathcal{L}_{\text{UTVAE-GEN}}(\theta; \phi) &= \mathcal{L}_{\text{UTVAE}}(\theta; \bar{\phi}) + \mathcal{L}_{\text{CEVAE}}(\phi; \bar{\theta})\\
    \mathcal{L}_{\text{UTVAE-INF}}(\theta; \phi) &= \mathcal{L}_{\text{CEVAE}}(\theta; \bar{\phi}) + \mathcal{L}_{\text{UTVAE}}(\phi; \bar{\theta})
\end{align*}
where $\bar{\theta}$ and $\bar{\phi}$ are fixed parameters - the gradients  with respect to these variables are blocked in the computational graph. 
We do so in order to isolate the impact of the choice of treatment distribution on the associated conditional distributions.
Thus, we get two separate objective functions for optimizing generative network parameters $\theta$ and inference network parameters $\phi$ with respect to fixing the other parameters $\bar{\phi}$ and $\bar{\theta}$ respectively.
Hence, the gradients of generative and inference networks are
\begin{align*}
    \nabla_{\theta,\phi} \mathcal{L}_{\text{RCTVAE-GEN}}(\theta; \phi) = \left[ \nabla_\theta \mathcal{L}_{\text{UTVAE}}(\theta; \bar{\phi}), \nabla_\phi \mathcal{L}_{\text{CEVAE}}(\phi; \bar{\theta}) \right]\\
    \nabla_{\theta,\phi} \mathcal{L}_{\text{RCTVAE-INF}}(\theta; \phi) = \left[ \nabla_\theta \mathcal{L}_{\text{CEVAE}}(\theta; \bar{\phi}), \nabla_\phi \mathcal{L}_{\text{UTVAE}}(\phi; \bar{\theta}) \right].
\end{align*}

\section{Classic causal inference and relationship to our proposed approach}
Our proposed method UTVAE address the discrepancy of estimated ATE between training and inference, which raises from having different treatment distribution in the original CEVAE. Another way to remove such a discrepancy is to rely on randomized controlled trial (RCT) without having a separate model \citep{Chalmers1981}. 
The treatments are randomly assigned to one of two groups, where one group is receiving the intervention that is being tested
and the other group is being controlled. The outcomes of two groups are then compared to measure the causal effect \citep{Hannan2008, Kendall164}.
RCT is deliberately designed to be unconfounded between the treatment $T$ and the confounder $Z$ which makes the study unbiased (see Figure~\ref{figs:graph_structure}b). 
The outcomes of these experiments do not suffer from the train and inference discrepancy as there is no training process.
However, it suffers in efficiency from running a large scale experiment.

An alternative way is to estimate conditional distributions from observational data and apply them for causal inference during inference time. 
A common classic approach is to re-weight the population outcomes using propensity score \citep{James1994, Hernan2006a, Breslow2009}.
That is, the expected outcome given a treatment is the expected observed outcome given the same treatment normalized by the propensity score, i.e., $\mathbb{E}[Y(t)] = \mathbb{E}\left[\frac{\mathbb{I}[T=t]Y(t)}{e(X)}\right]$, where the propensity score is defined as $e(X) = p(T=1|X=x)$. 
This effectively removes the edge between $X$ and $T$ as shown in Figure~\ref{figs:graph_structure}a.
The average treatment effect can be calculated using the expected outcomes as shown in the Appendix.
This method, known as {\em the inverse weighted probability}, became much more popular due to being able to deploy a large observational dataset \citep{Linden2016, Li2019, Bray2019}. 
UTVAE is closely related to the inverse weighted probability method except that the re-weighting is applied in training time. 
After training we can infer $\mathbb{E}[Y(1)]$ and $\mathbb{E}[Y(0)]$ using the trained model without further re-weighting.

More recent works estimates conditional distributions via learning representations that induces a balanced representation where the treated and control distributions are indistinguishable \citep{Johansson2016, Uri2017, Zhang2018, Zhang2020, Curth2021}. The learning objective usually consists of regularization term that minimize the representations of the factual and counterfactual distributions. 
For example, \cite{Johansson2016} propose to use IPM and Wasserstein distance to enforce domain invariance with distributional distances. Alternatively, because the model loses predictive power in domain invariant representations, \cite{Zhang2020} propose to enforce domain overlap in the posterior distribution of counterfactuals and add an invertible constraints to preserve the information content of the underlying context.  This is different from the approach of attempting to infer the joint distribution $p(X, Z)$ between the observation variables and the hidden confounders, and then using that knowledge to adjust for the hidden confounders \citep{Cai2012, Wang2016, Louizos2017, Pearl2021}. While these approaches identify the causal effect of $T$ on $Y$, our works is complementary in as sense that we can use uniform treatment distribution to train these latent variable causal models.
\section{Experiments}
We conducted our experiments with the questions of {\em is training a latent variable causal inference model using uniform treatment distribution any helpful}, and
similarly, {\em can we make use of both the uniform treatment and observation distribution to improve the performance of the model?}

The fundamental problem in causal inference is that the outcome of the treated and untreated events cannot be observed at the same time \citep{Saito2019}. 
It is impossible to check whether the prediction to a counterfactual question is correct in real life application.
The standard evaluation approach in research is to construct synthetic or semi-synthetic datasets, such that real data is adjusted in a way for us to know the true causal effect.
In our experiments, we follow the existing experiments in the literature \citep{Johansson2016, Louizos2017}.

We compare the four models: CEVAE, UTVAE, UTVAE-INF, and UTVAE-GEN on a synthetic \citep{Louizos2017} and IHDP dataset \citep{Hill2010} in order to answer the first two questions. \\
The {\bf synthetic dataset} is generated conditioned on the hidden confounder variable $Z$. Here is the generating process following the graph in Figure~\ref{figs:graph_structure}b:
\begin{align*}
    z_i\sim& \mathcal{B}(0.5)\\
    t_i|z_i \sim& \mathcal{B}(\alpha z_i+ (1-\alpha)(1-z_i))\\
    x_i|z_i \sim& \mathcal{N}(z_i, \rho^2_{z_1}+\rho^2_{z_0}(1-z_i))\\
    y_i|t_i,z_i \sim& \mathcal{B}(\sigma(3(z_i+2(2t_i-1)))),
\end{align*}
where the latent variable $Z$ is the mixture component, the treatment variable $T$ is a mixture of Bernoulli, the proxy to the confounder $X$ is a mixture of Gaussian distribution, the outcome $Y$ is determined as a Sigmoid function rate $\sigma$, and $\rho_{z_1}$ and $\rho_{z_0}$ are set to 3 and 5 respectively.
Following the previous experiment from \cite{Louizos2017}, proxy is 1-dimensional data, the confounder is 5-dimensional data points. The treatment assignment balance $\alpha$ is set to 0.75.
We generate 2000, 4000, 6000, and 8000 training data points in order to see the model performance with respect to the data size, and keep 1000 validation and
test data points for evaluation. We ran 30 cross-validations during the experiment. \\
The {\bf IHDP dataset} is a semi-simulated dataset based on the Infant Health and Development Program (IHDP).
IHDP conducted randomized experiments to discover the effect of high-quality child care and home visits for low-birth-weighted, premature infants.
The measured the future cognitive test scores of the treated children relative to controls \citep{Austin2008}.
The proxy has 25-dimensional covariates. \cite{Hill2010} removed subset of treated population in order to create a semi-simulated dataset.
The dataset consists of 747 data points with 138 treated and 608 control actions.
We repeat the experiments 8 times from artificially created imbalanced datasets from original IHDP dataset. The imbalance between treated and control subjects were generated by removing a subset of the treated population.

As mentioned before in the related work, $p(T|X)$ is known as a propensity score. The standard way to compute the propensity score is by fitting a logistic regression on treatment action $T$ given input $X$.
However, parametric models like logistic regression on an imbalance dataset tend to be overfit and lead to low generalization. We propose to use non-parametric methods instead such as kernel density estimation.
In our experiments, we use $\epsilon$-ball tree to compute the local density score of $p(T=1|X=x)$ and $p(T=0|X=x)$ \citep{Liu2006}. 

\begin{figure}[t]
    \centering
     \begin{minipage}[b]{0.24\textwidth}
         \includegraphics[width=\textwidth]{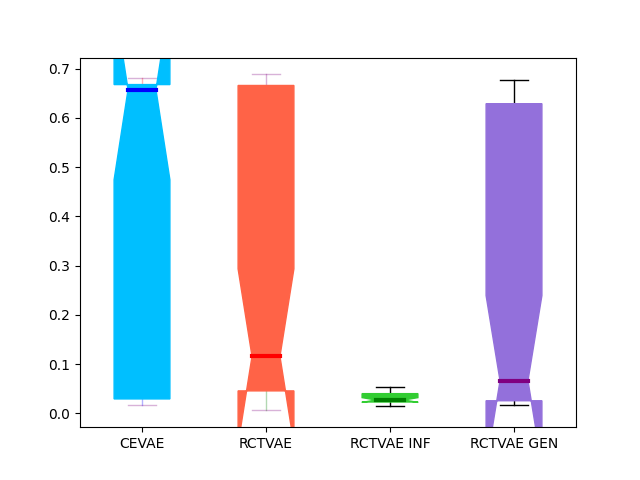}
              \vspace{-0.75cm}
         \caption*{2000 data points}
     \end{minipage}
     \hfill
     \begin{minipage}[b]{0.24\textwidth}
         \centering
         \includegraphics[width=\textwidth]{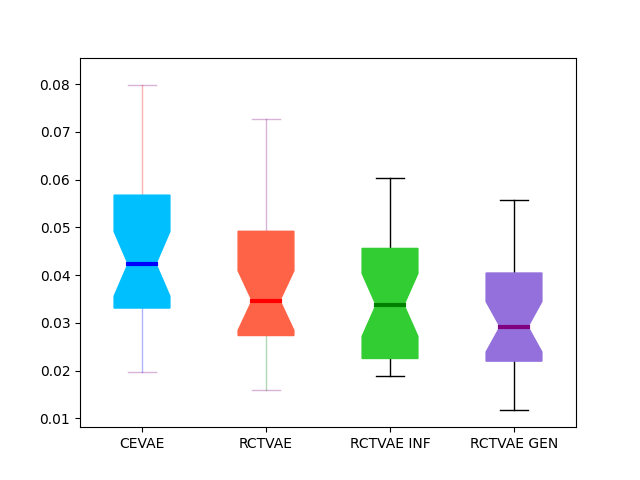}
              \vspace{-0.75cm}
         \caption*{4000 data points}
     \end{minipage}
     \hfill
     \begin{minipage}[b]{0.24\textwidth}
         \centering
         \includegraphics[width=\textwidth]{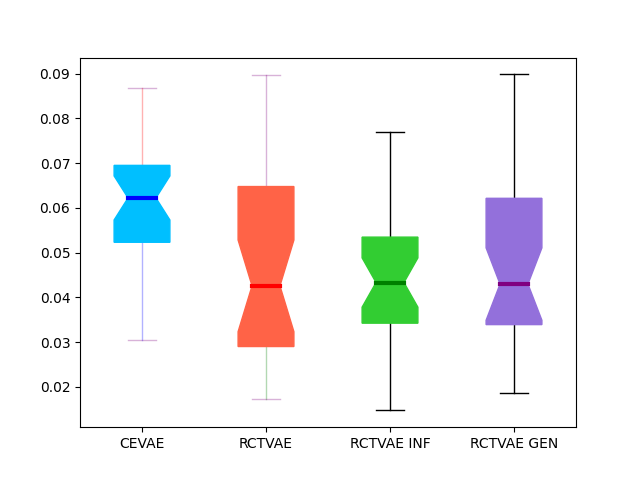}
              \vspace{-0.75cm}
         \caption*{6000 data points}
     \end{minipage}
     \hfill
     \begin{minipage}[b]{0.24\textwidth}
         \centering
         \includegraphics[width=\textwidth]{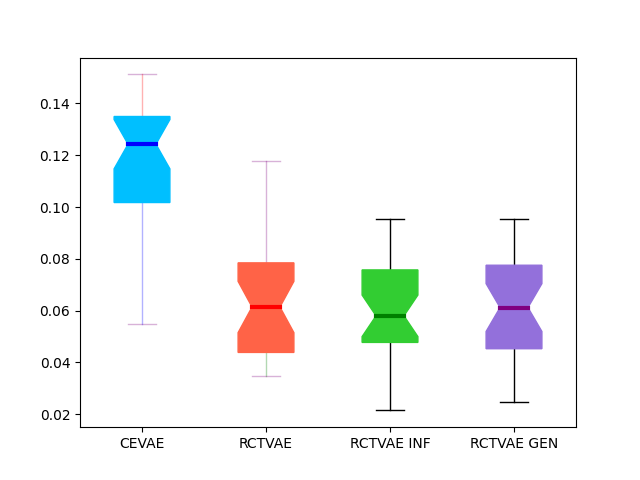}
        \vspace{-0.75cm}
         \caption*{8000 data points}
     \end{minipage}
    \caption{Mean absolute error between true and predicted ATEs with respect to various training set size on synthetic dataset.}
    \label{fig:syn_performance}
\end{figure}

\subsection{Model Evaluation}
Throughout the experiments, we follow the same experimental procedure and the same architecture for the generative and inference network from \cite{Louizos2017}.
Only the dimensionality of the proxy and the latent confounder layers differ, where we use 1 proxy and 5 latent dimensions for synthetic, and 25 input and 20 latent dimensions for IHDP dataset.
Because 19 variables were binary and 6 variables were continuous among 25 covariates, we use Bernoulli and Gaussian distribution for binary and continuous variables. 
We apply linear and softplus activations for mean and standard deviations of the Gaussian latent variables respectively. 
We run 100 and 200 epochs during the training for the two datasets.
We require one extra hyperparameter for training UTVAE that is the $\epsilon$ distance selection for measuring $p(T|X)$ using $\epsilon$-ball kernel density estimation.
We explore $\lbrace 0.5, 1, 1.5, 2 \rbrace$ and $\lbrace 2,2.5,3,3.5,4,5 \rbrace$ for synthetic and IHDP respectively. 

\begin{figure}[h]
    \centering
    \includegraphics[width=0.5\textwidth]{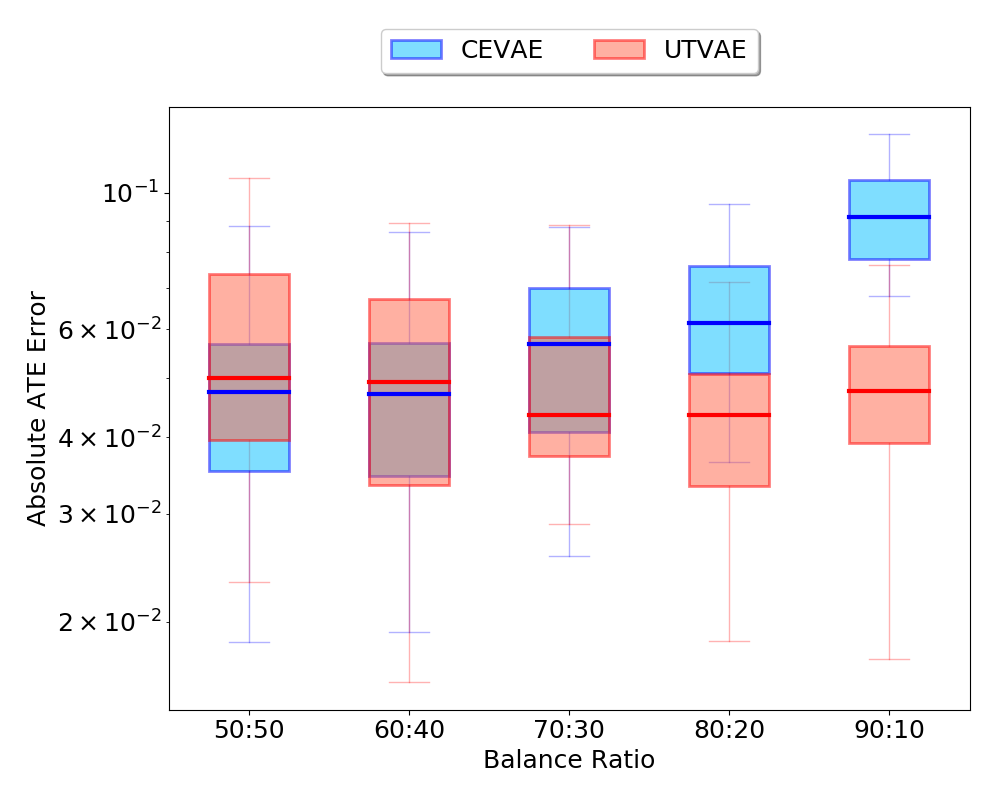}
    \vspace{-0.25cm}
    \caption{Absolute ATE error at different level of treatment assignment balance.}
    \label{fig:balance}
\end{figure}

By design UTVAE is ought to perform better than CEVAE when the treatment distribution is highly imbalanced. Likewise, CEVAE and UTVAE becomes identical as the treatment distribution becomes closer to the uniform distribution. Here, we verify our presumption by observing the absolute ATE error on synthetic datasets at multiple levels of treatment assignment balance. The synthetic datasets are generated using $\alpha=\lbrace 0.5, 0.6, 0.7, 0.8, 0.9 \rbrace$ where $\alpha$ decides the treatment assignment balance $t_i|z_i \sim \mathcal{B}(\alpha z_i+ (1-\alpha)(1-z_i))$.
Figure~\ref{fig:balance} presents the absolute ATE error for the two models. We observe that the error gap for CEVAE gets worse as the treatment labels become more imbalanced. In contrast, the error gap for UTVAE remains the same (or slightly) gets better with the imbalanced treatment labels. This suggests that UTVAE improves the performance on imbalanced datasets by applying importance weights to calibrate between observational and intervention distribution.

\begin{table}[h]
    \centering
    {\small 
    \begin{tabular}{c|c|c|c|c|c|c|c}
    \hline
          & RF  & BNN & CFRW & CEVAE & UTVAE & UTVAE-INF & UTVAE-GEN  \\\hline\hline
         ATE & 0.96     & 0.42  & 0.27 & 1.03  & 0.62  & .84  & 0.64  \\
         PEHE & 6.6     & 2.1   & 0.76 & 1.77  & 1.52  & 1.63 & 1.51  \\\hline
         \end{tabular}}
    \caption{The performance comparison - mean absolute ATE error and Precision in Estimation of Heterogeneous Effect (PEHE) error on IHDP dataset.}
    \label{tab:ihdp_performance}
    \vspace{-0.5cm}
\end{table}
\begin{figure}[h]
    \centering
     \begin{minipage}[b]{0.475\textwidth}
         \includegraphics[width=\textwidth]{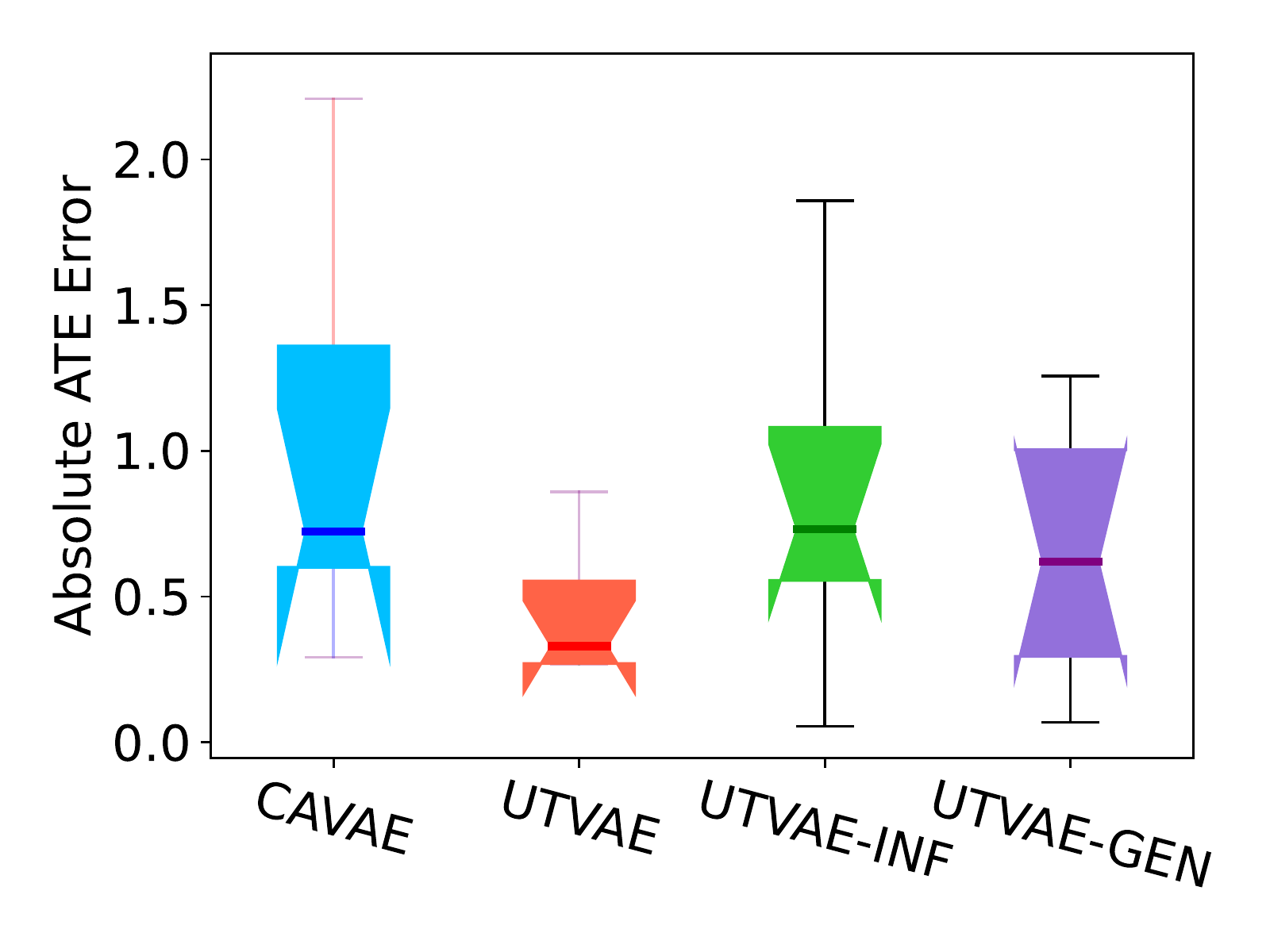}
              \vspace{-0.75cm}
         \caption*{Absolute ATE error}
     \end{minipage}
     \hfill
     \begin{minipage}[b]{0.475\textwidth}
         \centering
         \includegraphics[width=\textwidth]{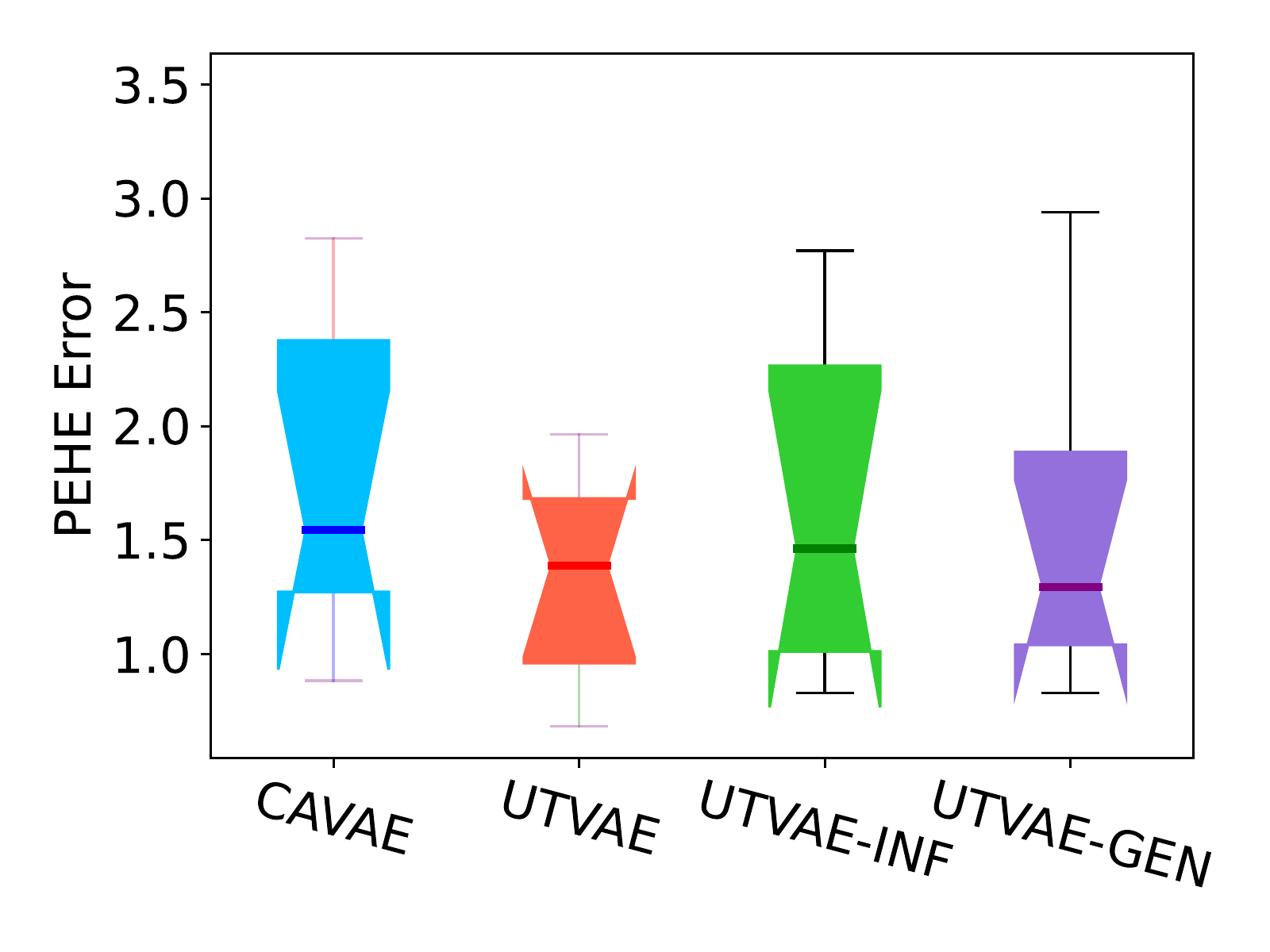}
              \vspace{-0.75cm}
         \caption*{PEHE error}
     \end{minipage}
    \caption{The performance comparison on IHDP dataset.}
    \label{fig:ihdp_performance}
    \vspace{0.5cm}
\end{figure}

We use mean absolute error between true ATE and predicted ATE. 
Previous work shows that CEVAE gets better absolute ATE error than logistic regression and TARnet \citep{Uri2017}. Here we show UTVAE performance compared to CEVAE.
The performance for the synthetic dataset is shown in Figure~\ref{fig:syn_performance}. 
The figure shows that UTVAE outperforms CEVAE, especially the performance gain increases as the number of samples increase.
Additionally, UTVAE-Gen tends to do better than UTVAE and UTVAE-Inf on average but not statistically significantly, when we have more than 2,000 data points. 
We notice that CEVAE, UTVAE, UTVAE-Gen have high ATE error rate for training with 2,000 data points, which indicates that 2,000 points are not enough for 
these models to perform well. In contrast, UTVAE-Inf gave consistent performance throughout using different numbers of data points. 

The mean absolute ATE and Precision in Estimation of Heterogeneous Effect (PEHE) performances for IHDP dataset are shown in Figure~\ref{fig:ihdp_performance}. 
Similar to above, we observe that all variants of UTVAE outperforms CEVAE. Both CEVAE and UTVAE-Inf have the highest variance in their performance. In contrast, UTVAE and UTVAE-Gen show tighter ATE and PEHE variance. UTVAE performs slightly better than UTVAE-Gen with ATE error metric, but it is vice versa with PEHE. From experimenting with both synthetic and IHDP dataset, it is clear that UTVAE and UTVAE-Gen give better results than CEVAE. This illustrates that training the generative models using uniform treatment samples or at least emulating with uniform treatment distribution improves the model performance. 


Table~\ref{tab:ihdp_performance} shows the performance against previously compared other methods as well. The ATE error results were taking directly from \cite{Louizos2017}. The Balancing Neural Networks \cite{Johansson2016} and Counterfactual Regression using Wasserstein distance \cite{Uri2017} denoted as  BNN and CFRW respectively. Both models emphasize learning {\em a balanced} representation between induced treated and control distribution. The difference is that they learn a deterministic representation while we learn a stochastic representation of the latent variable. We observe that BNN and CFRW performs better than CEVAE and UTVAE. It is  worth investigating what makes BNN and CFRW perform well.

\section{Conclusion}
Truly casual inference in practice remains an open problem in machine learning 
as the outcome of the treated and untreated events cannot be observed at the same time.
In this paper, we presented a novel causal inference algorithm, causal effect variational autoencoder with uniform treatment distribution (UTVAE), which
takes advantage of uniform  and observational treatment to mitigate the distribution shift that rises during test time. This procedure leads to ideal learning with better inference performance than the standard causal effect variational autoencoder (CEVAE). In the experiments, we empirically observed that the proposed UTVAE method can consistently help to improve the performance of CEVAE. Although we observed considerable improvements for our experiments with both synthetic and IHDP dataset, how to use our model in practice is still an open question. For the future work,
applying our UTVAE with a surrogate objective function that approximates ATE will be
important for more broad applications of causal inference.


\bibliographystyle{plain}
\bibliography{main}

\clearpage
\begin{appendix}

\appendix
\section{Supplementary Materials}

\subsection{Background: Inverse probability weighting}

Propensity score is the probability of taking treatment for a patient, $e(X) = p(T=1|X=x)$.
The famous propensity score theorem tells us that if we have unconfoundedness given $X$ and the positivity, then we also have unconfoundedness given $e(X)$ \citep{Imbens2015}:
\begin{align*}
(Y(1),Y(0)) \independent T | X) \implies (Y(1),Y(0)) \independent T | e(X)).
\end{align*}
The average treatment effect of $T$ on $Y$ is identified even with knowing $e(x)$ by the back-door criterion \citep{Pearl09a}.
Interestingly, this illustrates that the 1-dimension score function is enough to summarize the high-dimensional confounder $X$.\\

\noindent {\bf Pseudo population is unbiased estimate}\\
The difference in re-weighted population outcomes of treated and controlled actions are unbiased estimates of average treatment effect.
\begin{align*}
    \mathbb{E}[y|do(t=1)] - \mathbb{E}[y|\mathrm{do}(t=0)] = \mathbb{E}\left[\frac{\mathbb{I}[T=1]Y(1)}{e(X)}\right] - \mathbb{E}\left[\frac{\mathbb{I}[T=0]Y(0)}{1-e(X)}\right],
\end{align*}
We can see that $\mathbb{E}\left[\frac{Y(1)T}{e(X)}\right] = \mathbb{E}\left[ Y(1) \right]$ because
\begin{align*}
    \mathbb{E}\left[\frac{Y(1)T}{e(X)}\right] &= \mathbb{E}\left[ \mathbb{E} \left[ \frac{Y(1)T}{p(T|X)} | X \right] \right] \\
        &= \mathbb{E}\left[ \frac{ \mathbb{E} \left[ Y(1) | X \right]  \mathbb{E} \left[ T | X \right]}  {p(T|X)} \right] \\
        &= \mathbb{E}\left[ \mathbb{E} \left[ Y(1)|X)\right]\right]\\
        &= \mathbb{E}\left[ Y(1) \right]
\end{align*}
and similar derivation applies for $Y(0)$ case.
\end{appendix}

\subsection{Covid-19 Treatment Causal Effect Analysis}

{\bf Full list of Covid-19 EHR patient features}\\
    Age, BodyMassIndex, white, black, asian,
    hispanic, hypertension, afib, valve disease,
    asthma, psych, Myocardial infarction,
    Congestive heart failure, Peripheral vascular disease,
    Cerebrovascular disease, Dementia,
    Chronic obstructive pulmonary disease, Rheumatoid disease,
    Peptic ulcer disease, Mild liver disease,
    Diabetes without chronic complications,
    Diabetes with chronic complications, Hemiplegia or paraplegia,
    Renal disease, Cancer (any malignancy),
    Moderate or severe liver disease, Metastatic solid tumour,
    AIDS/HIV, Charlson score, Weighted Charlson score, Sex Female,
    Sex Male, Smoking Level 0, Smoking Level 1, Smoking Level 2,
    Smoking Level 3, Smoking Level 4, Smoking Level 5, Smoking Level 6,
    Smoking Level 7, Smoking Level 8, Ethnicity Unspecified,
    Ethnicity Not of Spanish/Hispanic Origin, Ethnicity Patient Refused,
    Ethnicity Spanish/Hispanic Origin, Ethnicity Unknown,
    Charlson index 0, Charlson index 1-2, Charlson index 3-4,
    Charlson index $\geq$ 5, Weighted Charlson index 0,
    Weighted Charlson index 1-2, Weighted Charlson index 3-4,
    Weighted Charlson index $\geq$ 5

\begin{figure}[t]
\centering
\vspace{-0.5cm}
\begin{minipage}[b]{0.49\textwidth}
\begin{minipage}[b]{0.475\textwidth}
    \includegraphics[width=\linewidth]{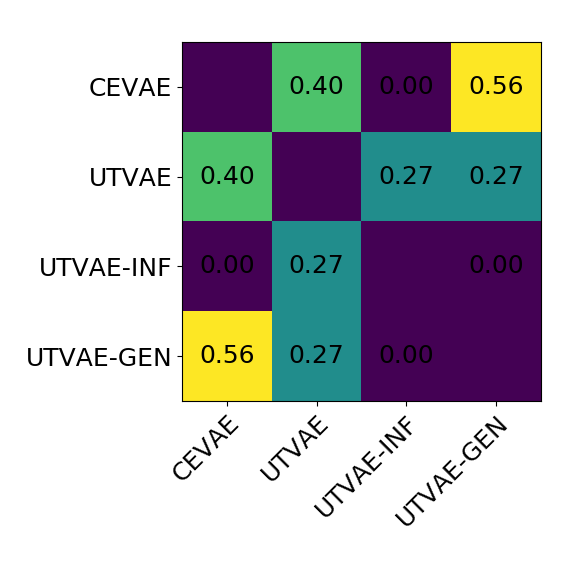}
    \vspace{-1.0cm}
    \caption*{Recovery}
\end{minipage}
\begin{minipage}[b]{0.425\textwidth}
    \vspace{-0.5cm}
    \includegraphics[width=\linewidth]{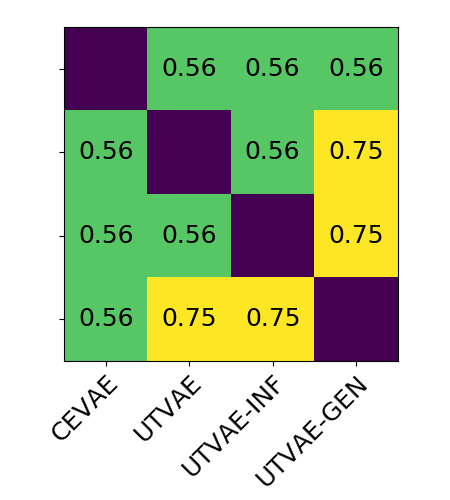}
    \vspace{-1.0cm}
    \caption*{Worsen}
\end{minipage}
\vspace{-0.25cm}
\caption*{Among top 7 Treatments}
\end{minipage}
\begin{minipage}[b]{0.495\textwidth}
\begin{minipage}[b]{0.475\textwidth}
    \includegraphics[width=\linewidth]{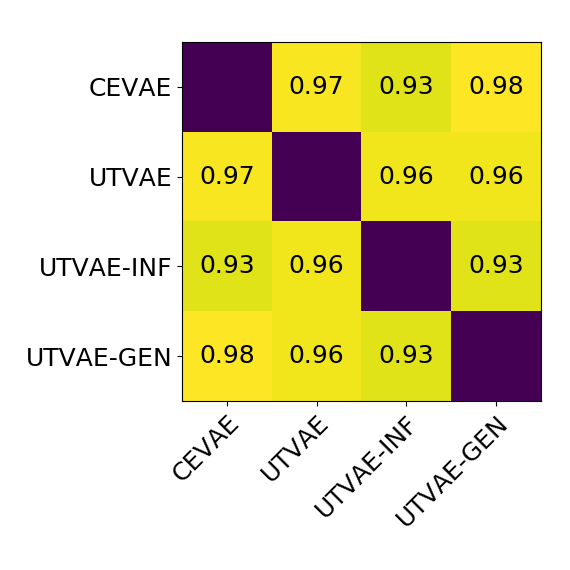}
    \vspace{-1.0cm}
    \caption*{Recovery}
\end{minipage}
\begin{minipage}[b]{0.475\textwidth}
    \vspace{-0.5cm}
    \includegraphics[width=\linewidth]{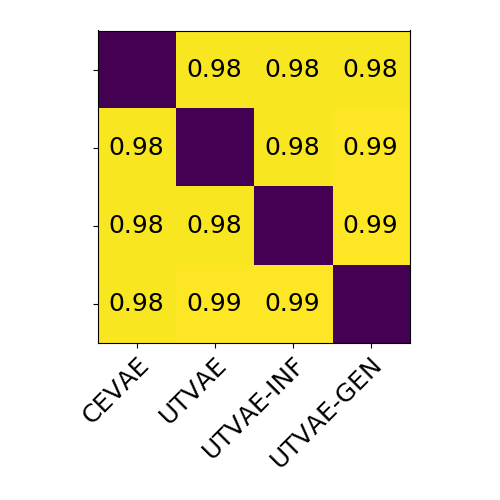}
    \vspace{-1.0cm}
    \caption*{Worsen}
\end{minipage}
\vspace{-0.25cm}
\caption*{Among bottom 7 Treatments}
\end{minipage}
\vspace{-0.25cm}
\caption{Intersection over Union on top 7 treatments among CEVAE, UTVAE, UTVAE-ING, and UTVAE-GEN.}
\label{figs:iou_pairwise}
\end{figure}

\subsection{Covid-19 Treatment Causal Effect Analysis}
In this section, we apply our method to one of the challenging and important problems in today's world - the Covid19 treatment discovery.
Our goal is to analyze the effects of various Covid-19 treatments on patients' recovery and hope to take a step towards understanding the relationship between
the Covid-19 virus and the treatments that were used during the pandemics. 

Here, we use electronic health records (EHR) data from NYU Langone Health that were gathered from March 2019 to June 2019.
The EHR data consists of 16,978 patients from New York and the data has 54 types of patient features that contains
basic patient features like age, body mass index, sex, and smoking level, as well as the previous patient disease records like
whether they have/had asthma, chronic obstructive pulmonary disease, cerebrovascular disease, myocardial infarction, and so on.
The full list of features are included in the appendix. 
There are a total of 2883 treatments among medication name and pharmaceutical class groups. We focused on 445 treatments that fall under pharmaceutical class. 
We have five indicators which can be used as potential outcomes of treatments, which are 
Covid-19 flag, inpatient flag, adverse event ICU transfer, adverse event intubated, and adverse event mortality.

\begin{wrapfigure}{r}{0.5\textwidth}
    \centering
     \begin{minipage}[b]{0.245\textwidth}
         \includegraphics[width=\textwidth]{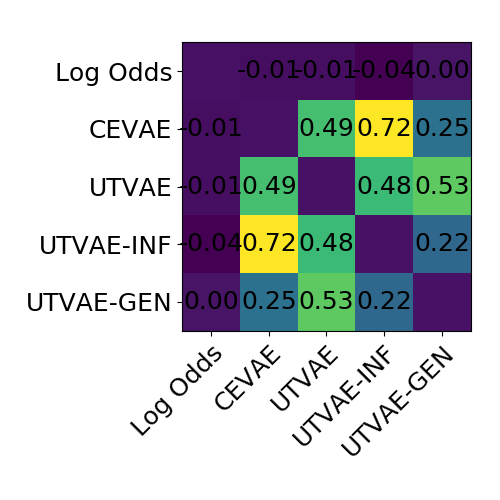}
              \vspace{-0.75cm}
         \caption*{Recovery}
     \end{minipage}
     \hfill
     \begin{minipage}[b]{0.245\textwidth}
         \centering
         \includegraphics[width=\textwidth]{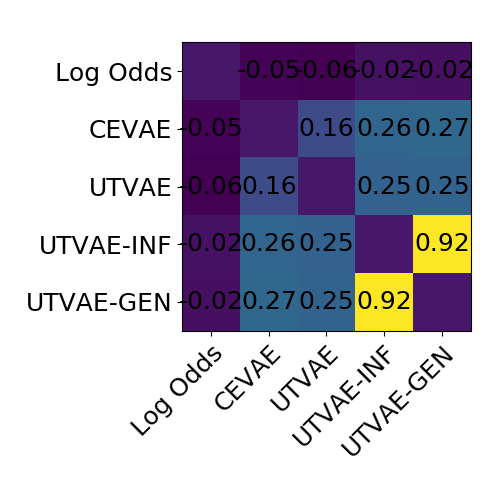}
              \vspace{-0.75cm}
         \caption*{Severe}
     \end{minipage}
    \caption{Kendall’s tau correlation among adjusted log-odds, CEVAE, UTVAE, UTVAE-INF, and UTVAE-GEN shown. - Kendall’s tau correlation is a measure of the correspondence between two rankings.}
    \label{fig:covid_rankcorrelation}
    \vspace{0.5cm}
\end{wrapfigure}

\begin{figure}[t!]
    \begin{minipage}[b]{\textwidth}
    \centering
    \includegraphics[width=\linewidth]{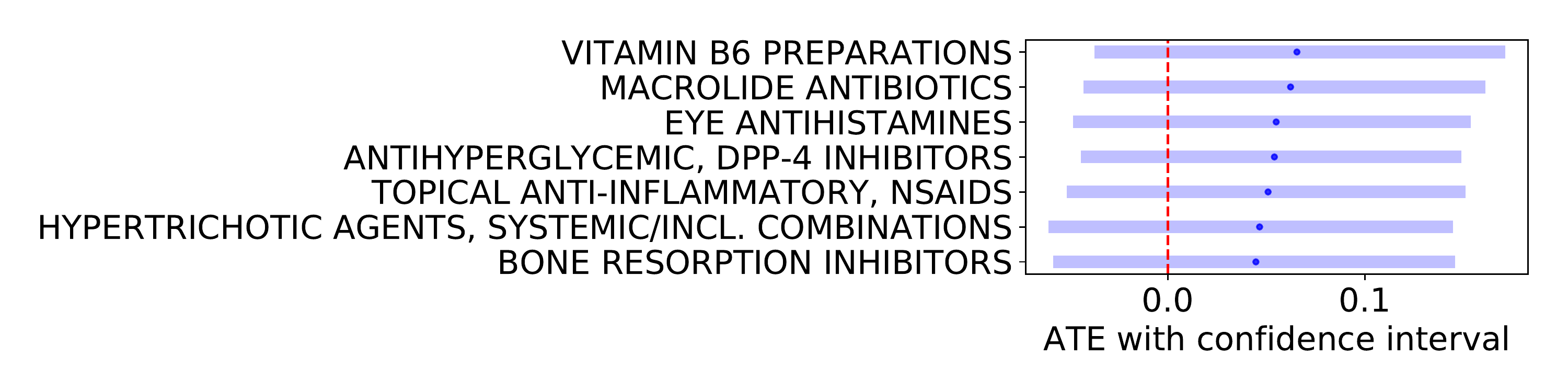}
    \vspace{-1cm}
    \caption*{The treatment with top 20 ATEs that are trained from recovery dataset}
    \end{minipage}\\%
    \begin{minipage}[b]{\textwidth}
    \centering
    \includegraphics[width=\linewidth]{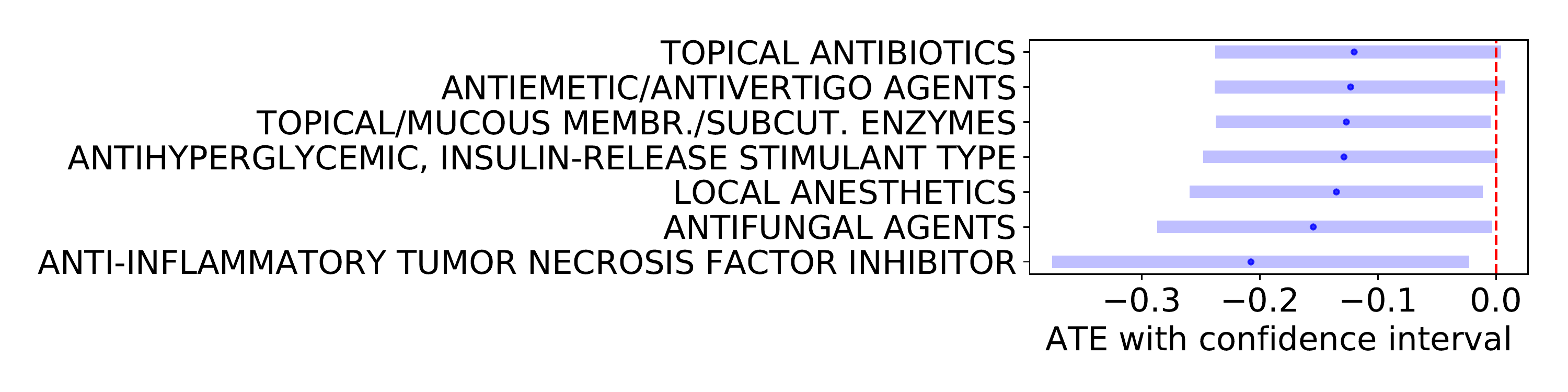}
    \vspace{-1cm}
    \caption*{The treatment with bottom 20 ATEs that are trained from recovery dataset}
    \end{minipage}%
    \caption{The top and bottom 20 ATEs over different treatments with confidence interval on recovery dataset - the model was trained using UTVAE.}
    \vspace{1.0cm}
    \begin{minipage}[b]{\textwidth}
    \centering
    \includegraphics[width=\linewidth]{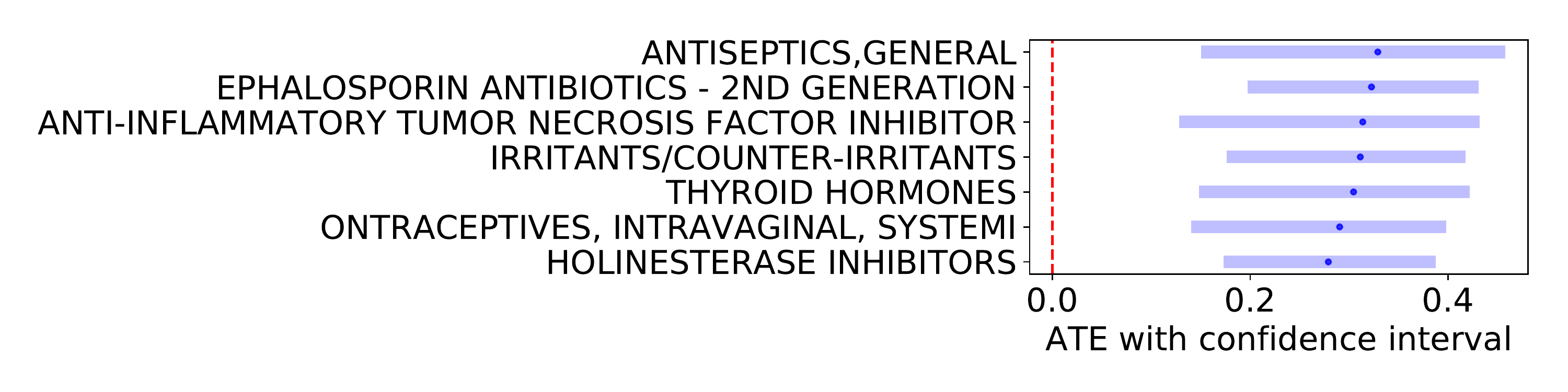}
    \vspace{-1cm}
    \caption*{The treatment with top 20 ATEs that are trained from worsen dataset}
    \end{minipage}\\%
    \begin{minipage}[b]{\textwidth}
    \centering
    \includegraphics[width=\linewidth]{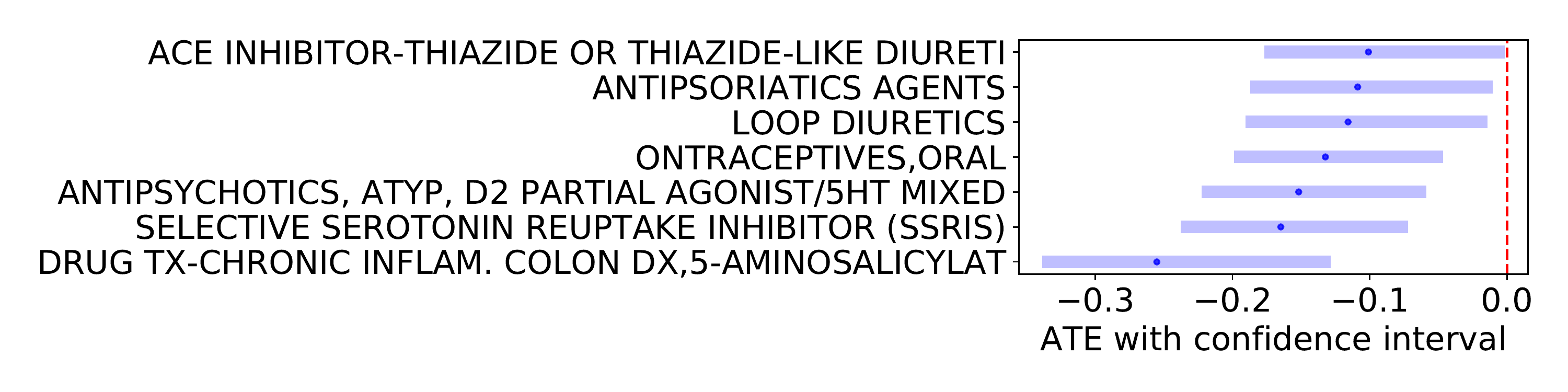}
    \vspace{-1cm}
    \caption*{The treatment with bottom 20 ATEs that are trained from worsen dataset}
    \end{minipage}%
 \caption{The top and bottom 20 ATEs over different treatments with confidence interval on worsen dataset - the model was trained using UTVAE.}
 \label{figs:covid_ate_exp1}
\end{figure}

We conduct two experiments by asking {\em which treatments help} and {\em which treatments make worse?}
In order to answer the first question, we construct the recovery label simply by using Covid-19 flag.
We call this {\em Recovery dataset}.
The Covid-19 flag indicates whether the test results in negative after the treatment or not.
We call this {\em Worsen dataset}.
In order to answer the latter question, we construct the adverse label by checking whether
the patient was transferred to ICU, inbutated, or died. 
We train a causal model as a binary treatment actions one at a time out of 445 treatments.
For both experiments, we normalized the data between zero mean and one standard deviation, and partition the dataset into 70\% training and 30\% validation set.
We used Adam for optimization with a 0.00007 learning rate and ran 220 iterations.
Among the 445 causal models, we eliminate the models that are trained with 20 or less number of treated cases or zero treated cases among treated patients.

We first check the proxy $X$ is helpful for discovering confounder by comparing against the empirical probability that is marginalized over $X$, $p(Y|T)$ and compare against our trained models. We compute Kendall’s tau correlation

Let us first check the treatments that agrees between the four models.
We consider the top 7 treatments of CEVAE, UTVAE, UTVAE-INF, and UTVAE-GEN based on ATE and compute the intersection over union (IOU).
Figure~\ref{figs:iou_pairwise} presents the IOU score between pairs of four models for both recovery and worsen dataset.
We observe that there are less overlaps between models for recovery dataset compare to worsen dataset.
The IOU rate of the four models is 0\% but we get 25\% IOU excluding UTVAE-INF on recovery dataset and the common treatments are 'vitamin B6 preparations', 'macrolide antibiotics', 'topical anti-inflammatory, NSAIDS'.
The IOU rate is 50\% excluding UTVAE-INF on worsen dataset and the common treatments are 'anti-inflammatory tumor necrosis factor inhibitor', 'irritants/counter-irritants', 'antiseptics, general', 'cephalosporin antibiotics - 2nd generation', and 'contraceptives intravaginal, systemic'.
The results illustrate that UTVAE-INF does not tend to agree with other three models.

When we consider the 95\% confidence interval of ATE on top 7 and bottom 7 treatments of UTVAE in Figure~\ref{figs:covid_ate_exp1}, we find that the confidence intervals cross the origin for top 7 and bottom 7 treatments that are learned from the recovery dataset. In contrast, the confidence interval does not cross the origin for top 7 and bottom 7 treatments that are learned from the worsen dataset.
It is ideal that these intervals do not cross the origin, because this implies that the treatment results is consistent 95\% of times.
As we are working with scarcity of treated labels and its outcome labels, we speculate that it is easier to answer whether the treatment is not working than if it is working. From this experiments, we find that understanding which Covid-19 treatment is effective is inclusive, because the recovery dataset is scarce, noisy, and the problem itself is very hard. 

\end{document}